\documentclass{pf2003}

\usepackage{verbatim}
\usepackage[latin1]{inputenc}
\usepackage[T1]{fontenc}
\usepackage{url} 
\usepackage{color} 

\usepackage{amsfonts}
\usepackage{amssymb}
\usepackage{amsmath}
\usepackage[dvips]{graphicx}

\newenvironment{proof}{\list{}{\parsep 0em \leftmargin 2em \rightmargin 0em}%
                \item\relax \small Preuve : }{\hfill $\square$ \endlist}

\newtheorem{definition}{Définition}
\newtheorem{example}{Exemple}
\newtheorem{proposition}{Proposition}
\newtheorem{property}{Propriété}

\DeclareMathOperator{\comp}{\theta}

\newcommand{\txtrm}[1]{\textup{\sffamily #1}}
\newcommand{\textalgo}[1]{\txtrm{#1}}
\newcommand{\textagg}[1]{\text{\small #1}}

\newcommand{\Charm}{\textalgo{CHARM}}
\newcommand{\Closet}{\textalgo{CLOSET}}
\newcommand{\Carpenter}{\textalgo{CARPENTER}}
\newcommand{\Dminer}{\textalgo{D-Miner}}

\newcommand{\SOM}{\textagg{SOM}}
\newcommand{\MIN}{\textagg{MIN}}
\newcommand{\MAX}{\textagg{MAX}}
\newcommand{\MOY}{\textagg{MOY}}

\newcommand{\Concept}{\txtrm{Concepts}} 
\newcommand{\Ferme}{\txtrm{Fermés}} 

\newcommand{\BB}{\ensuremath{\mathcal{B}}}
\newcommand{\OO}{\ensuremath{\mathcal{O}}}
\renewcommand{\AA}{\ensuremath{\mathcal{A}}}
\newcommand{\db}{\ensuremath{bd}}
\newcommand{\freq}{\ensuremath{\txtrm{Freq}}}

\newcommand{\CC}{\ensuremath{\mathcal{C}}}

\newcommand{\proj}[1]{{p(#1)}}
\newcommand{\pCC}{{\proj \CC}}

\newcommand{\cl}{\txtrm{cl}}

\newcommand{\Cfreq}{\CC_{\gamma\txtrm{-freq}}}
\newcommand{\Csous}[1]{\CC_{\sseq {#1}}}
\newcommand{\Csur}[1]{\CC_{\supseteq {#1}}}

\newcommand{\supalpha}{\txtrm{sup}_\alpha}

\newcommand{\et}{\wedge}
\newcommand{\ou}{\vee}
\newcommand{\equ}{\Leftrightarrow}

\newcommand{\set}[1]{\left\{ #1 \right\}}
\newcommand{\size}[1]{\left| #1 \right|}

\newcommand{\sseq}{\subseteq}

\newcommand{\ra}{\,r_a\,}
\newcommand{\ro}{\,r_o\,}

\newcommand{\at}[1]{a_{#1}}
\newcommand{\ob}[1]{o_{#1}}

\newcommand{\defnt}[1]{\textbf{#1}}

\shorttitle{Extraction de concepts sous contraintes}

\shortouvrage{CAp 2005}

\title{Extraction de concepts sous contraintes dans des données d'expression de gènes\thanks{Ce travail
a été partiellement financé par l'ACI masse de données (MD 46, Bingo)}}

\author{Baptiste Jeudy\inst{1}, François Rioult\inst{2}}

\institute{
Équipe Universitaire de Recherche en Informatique de St-Etienne (EURISE),\\
Université de St-Etienne.\\
\texttt{baptiste.jeudy@univ-st-etienne.fr}
\and
GREYC - CNRS UMR 6072,\\
Université de Caen Basse-Normandie\\
\texttt{francois.rioult@info.unicaen.fr}
}

\begin{document}
\maketitle

\begin{abstract}
  L'une des activités les plus importantes en biologie est l'analyse des données d'expression de
  gènes. Les biologistes espèrent ainsi mieux comprendre les fonctions des gènes et leurs
  interactions.  Nous étudions dans cet article une technique permettant d'aider à l'analyse de ces
  données d'expression : l'extraction de concepts sous contraintes. Pour cela, nous proposons
  d'extraire des fermés sous contraintes dans les données ``transposées'' en utilisant des
  algorithmes classiques. Ceci nous amène a étudier la ``transposition'' des contraintes dans les
  données transposées de manière à pouvoir les utiliser dans ces algorithmes.

  \motscles{Extraction de connaissances, Data-mining, Concepts Formels, Itemsets Fermés,
    Contraintes.} 
\end{abstract}

\section{Motivations}

Maintenant que le décodage du génome est terminé pour de nombreuses espèces animales et végétales,
il reste encore un formidable défi pour la biologie moderne : comprendre la fonction de tous ces
gènes et la manière dont ils interagissent entre-eux.  Pour cela, les biologistes mènent des
expériences de mesure de l'expression de gènes.  Celles-ci ont pour but de leur fournir des données
leur permettant de faire des hypothèses sur ces fonctions et ces interactions.

Les données d'expression de gènes se présentent typiquement sous la forme d'une matrice binaire.
Chaque colonne représente un gène et chaque ligne donne les résultats d'une expérience de mesure du
niveau d'expression des gènes.  Chacune de ces expériences consiste à déterminer, pour une cellule
donnée issue d'une situation biologique donnée (par exemple un organe spécifique, une culture
cellulaire), quels sont les gènes qui sont sur-exprimés, c'est-à-dire ceux qui ont une activité
biologique importante au moment de la mesure.  Dans la matrice, les gènes qui sont
sur-exprimés\footnote{dont l'activité biologique dépasse un seuil fixé par le biologiste} dans une
situation biologique sont codés par un 1. Ceux qui ne le sont pas sont codés par un 0. La
table~\ref{tab:mat} donne un exemple d'une telle matrice.

\begin{table}[h]
  \centering
  \begin{tabular}{c|cccc}
               & Gène 1 & Gène 2 & Gène 3 &Gène 4 \\\hline
     cellule 1 & 1 & 1 & 1 & 0 \\
     cellule 2 & 1 & 1 & 1 & 0 \\
     cellule 3 & 0 & 1 & 1 & 1 \\
  \end{tabular}
  \caption{Exemple de matrice d'expression de gènes}
  \label{tab:mat}
\end{table}

Dans cet article, nous étudions une technique de fouille de données permettant d'aider le biologiste
à faire des hypothèses sur les fonctions des gènes et la manière dont ils interagissent. Pour cela,
les techniques d'extraction de motifs semblent particulièrement adaptées. Il existe cependant de
nombreux types de motifs : les itemsets, les itemsets fermés ou libres, les règles
d'association ou encore les concepts formels. Nous avons choisi ici d'étudier l'extraction des concepts.

Dans ce cadre, un concept formel est une paire $(G,E)$ où $G$ est un ensemble de gènes (i.e., un ensemble
de colonnes de la matrice) appelé intension du concept et $E$ un ensemble d'expériences (i.e., un
ensemble de lignes) appelé extension du concept. Ces ensembles sont tels que si $g \in G$ et $e \in
E$, alors le gène $g$ est sur-exprimé dans l'expérience $e$ (il y a un 1 dans la ligne $e$ colonne
$g$). De plus, les deux ensembles $G$ et $E$ sont maximaux, i.e., ils ne peuvent pas grossir sans
perdre la propriété précédente (une définition plus formelle des concepts est donnée dans la
section~\ref{sec:definitions}). Autrement dit, un concept est une sous-matrice maximale ne contenant
que des 1.  Dans notre matrice exemple, (\{Gène 1, Gène 2, Gène 3\}, \{cel 1, cel 2 \}) est un
concept. 

Du point de vue du biologiste, les concepts sont très intéressants.  En effet, un concept $(G,E)$
regroupe des gènes qui sont sur-exprimés dans les mêmes expériences. Si la fonction de certains de
ces gènes est connue, cela peut permettre de faire des hypothèses sur la fonction de ceux qui sont
inconnus. De plus, si les expériences apparaissant dans l'extension $E$ partagent des propriétés
communes (par exemple, elles concernent toutes des cellules du foie ou des cellules cancéreuses),
cela permet encore une fois de faire des hypothèses sur les gènes.  Le fait que les concepts
associent à la fois des gènes et des expériences est donc un avantage par rapport à d'autres motifs
comme les itemsets ou les règles d'association qui ne portent que sur les gènes.  De plus, un gène
(ou une expérience) peut apparaître dans plusieurs concepts (par opposition à ce qui se passe dans
le cas du clustering). Si le biologiste s'intéresse à un gène particulier, il peut donc étudier
quels sont les gènes liés à celui-ci (i.e., apparaissant dans les mêmes concepts) suivant les
situations biologiques. Cela est très important car il s'avère en effet qu'un gène peut intervenir
dans plusieurs fonctions biologiques différentes.  Enfin, les concepts sont beaucoup moins nombreux
que les itemsets tout en représentant la même information~: ils sont donc plus simples à exploiter.

Pour simplifier encore l'exploitation de ces concepts par le biologiste, l'utilisation de
contraintes semble pertinente~: le biologiste peut indiquer une contrainte qui doit être satisfaite
par tous les concepts extraits. Par exemple, il peut imposer qu'un gène particulier (ou ensemble de
gènes) apparaisse (ou pas) dans les concepts extraits. Il peut aussi se restreindre aux concepts
impliquant des expériences sur des cellules cancéreuses ou contenant au moins 5 gènes.
L'utilisation des contraintes permet finalement au biologiste de mieux cibler sa recherche.

\subsection{Notre contribution}
\label{sec:contrib}

\begin{sloppypar}
Nous proposons dans cet article d'étudier l'extraction de concepts sous contraintes dans des
données d'expression de gènes. Cette extraction pose deux problèmes principaux :
\end{sloppypar}

\begin{enumerate}
\item utilisation des contraintes~: nous laissons la possibilité à l'utilisateur
  de spécifier une contrainte portant à la fois sur l'intension et l'extension du concept. 
  Ces contraintes sont utiles pour l'utilisateur pour préciser sa recherche mais elles sont
  aussi parfois indispensables pour rendre l'extraction faisable. En effet, il est généralement
  impossible d'extraire tous les concepts. Il faut donc dans ce cas utiliser les contraintes
  {\em pendant} l'extraction (et non pas seulement dans une phase de filtrage des concepts {\em après} 
  l'extraction) pour diminuer la complexité celle-ci.

\item taille des données~:
  la complexité des algorithmes d'extraction est généralement linéaire par rapport au nombre de lignes
  et exponentielle par rapport au nombre de colonnes. 
  Or dans le cas des données d'expression de gènes, le nombre de colonnes est souvent très
  important : l'utilisation de techniques comme les puces à ADN permet d'obtenir l'expression de
  milliers de gènes en une seule expérience. D'un autre coté, le nombre d'expériences est souvent
  réduit du fait du temps nécessaire à leur mise en place et de leur coût.  Ceci amène à des
  matrices comportant beaucoup de colonnes (jusqu'à plusieurs milliers) et relativement peu de
  lignes (quelques dizaines ou centaines) ce qui est plutôt atypique dans le domaine du data-mining. 
  Les algorithmes classiques ne sont donc pas bien adaptés à ce type de données. 

\end{enumerate}

L'extraction de motifs sous contrainte est un thème de recherche qui a été très étudié ces dernières
années~\cite{srikantetal97,ngetal98,garofalakisetal99,boulicautjeudy00a,hankdd00,zaki00b,boulicautjeudy01a,bucila03,hunor03,bonchi03,bonchi04a}...
De nombreux algorithmes ont été proposés et tentent d'utiliser
efficacement les contraintes pour diminuer les temps d'extraction en élaguant le plus tôt possible
l'espace de recherche.  L'extraction de concepts est fortement liée à l'extraction d'itemsets libres
ou fermés dont l'étude a également donné lieu à de nombreux
travaux~\cite{pasquieretal99,boulicautetal00c,peietal00b,zaki02,bbr03}...

Cependant, ces travaux ne font pas d'extraction de concepts sous contrainte et ne sont pas adaptés 
à des données ayant plus de colonnes que de lignes. En ce qui concerne l'extraction de concepts
sous contraintes, une proposition récente à été faite dans~\cite{brb04}. Cependant, l'algorithme
proposé, \Dminer, ne permet que de traiter un type particulier de contraintes, les contraintes
monotones. Nous verrons dans la section~\ref{sec:util_proj} comment l'étude que nous proposons ici
va nous permettre de traiter aussi les contraintes anti-monotones avec cet algorithme.

En ce qui concerne le second problème, 
plusieurs propositions ont été faites récemment pour le résoudre : l'algorithme
\Carpenter~\cite{pct03} est conçu pour extraire les fermés fréquents dans une base de données avec
plus de colonnes que de lignes.  Dans~\cite{rbc03,RC031}, les auteurs utilisent des
algorithmes classiques mais au lieu de faire l'extraction dans les données originales, ils
travaillent sur la matrice transposée. Dans ce cas, la matrice transposée comporte beaucoup de
lignes et peu de colonnes, ce qui permet d'utiliser les techniques habituelles efficacement. 
Cependant, ces travaux ne traitent que du cas de la contrainte de fréquence ou de contraintes
simples sur les itemsets. Le cas général où la contrainte est une formule booléenne 
construite à partir de contraintes simples, portant à la fois sur l'intension et l'extension, 
n'est pas abordé.

Notre proposition est donc d'utiliser des algorithmes classiques (éventuellement légèrement modifiés) 
dans la matrice transposée, afin de travailler sur des données au format
plus classique (peu de colonnes, beaucoup de lignes).  Pour pouvoir traiter des
contraintes complexes portant sur les concepts, nous allons présenter ici une étude théorique
sur les contraintes et sur la manière de les ``transposer'' (en fait, il s'agira plutôt d'une
projection) de façon à pouvoir les utiliser dans la matrice transposée. 

Cet article est organisé de la manière suivante : dans la section~\ref{sec:definitions},
nous rappelons quelques définitions à propos de l'extraction d'itemsets et de la correspondance
de Galois. Nous présentons ensuite formellement le problème que nous cherchons à résoudre. 
Dans la section~\ref{sec:proj}, nous présentons la projection des contraintes
simples et composées. Ensuite, la section~\ref{sec:util_proj} montre comment utiliser
la projection de contraintes et l'extraction dans la matrice transposée pour 
résoudre notre problème. Finalement, nous concluons dans la section~\ref{sec:conclusion}.

\section{Définitions}
\label{sec:definitions}

Pour éviter les confusions entre les lignes (ou colonnes) de la base de données originale et les
lignes (ou colonnes) de base de données ``transposée'', nous définissons une base de données comme
une relation entre deux ensembles~: un ensemble d'attributs et un ensemble d'objets. L'ensemble des
\defnt{attributs} (ou items) est noté $\AA$ et correspond, dans notre application biologique, à
l'ensemble des gènes. L'ensemble des \defnt{objets} est noté $\OO$ et représente les situations
biologiques. L'\defnt{espace des attributs}, $2^\AA$, est la collection des sous-ensembles de $\AA$,
appelés \defnt{itemsets} et l'\defnt{espace des objets}, $2^\OO$, est
la collection des sous-ensembles de $\OO$. 
Lorsqu'on considère l'ordre défini par l'inclusion ensembliste, chacun des espaces $2^\AA$ et 
$2^\OO$ est naturellement muni d'une structure de treillis. 

Une base de données est une relation binaire de $\AA \times \OO$ et peut être représentée par une
matrice booléenne dont les colonnes sont les attributs et les lignes sont les objets. Cette matrice
constitue la représentation originale de la base. Au cours de cet article, nous considérerons
que la base de données a plus d'attributs que d'objets et nous utiliserons également la
représentation transposée des données, où les attributs de la base sont portés sur les lignes et
les objets sur les colonnes (cf. Table~\ref{tab:database}).

\begin{table}
  \hfill
  \begin{tabular}{c|cccc}
              & $\at{1}$ & $\at{2}$ & $\at{3}$ &$\at{4}$ \\\hline
     $\ob{1}$ & 1 & 1 & 1 & 0 \\
     $\ob{2}$ & 1 & 1 & 1 & 0 \\
     $\ob{3}$ & 0 & 1 & 1 & 1 \\
  \end{tabular}
  \hfill
  \begin{tabular}{c|ccc }
              & $\ob{1}$ & $\ob{2}$ & $\ob{3}$  \\\hline
     $\at{1}$ & 1 & 1 & 0 \\
     $\at{2}$ & 1 & 1 & 1 \\
     $\at{3}$ & 1 & 1 & 1 \\
     $\at{4}$ & 0 & 0 & 1 \\
  \end{tabular}
  \hspace*{\fill}
  \caption{Représentation originale et transposée de la base de données
  présentée table~\ref{tab:mat}. Les attributs sont $\AA = \set{\at{1},\at{2},\at{3},\at{4}}$ et les
  objets sont $\OO=\set{\ob{1},\ob{2},\ob{3}}$. Nous utilisons une notation sous forme de chaîne
  pour les ensembles, par exemple $\at{1}\at{3}\at{4}$ désigne l'ensemble d'attributs
  $\set{\at{1},\at{3},\at{4}}$ et $\ob{2}\ob{3}$ désigne l'ensemble d'objets $\set{\ob{2},\ob{3}}$. 
  Cette base de données sera utilisée dans tous les exemples.} 
  \label{tab:database}
\end{table}

\subsection{Correspondance de Galois}
\label{sec:galois}

L'idée principale qui fonde notre travail est d'utiliser la correspondance forte entre les treillis des
$2^\AA$ et $2^\OO$, appelée \defnt{correspondance de Galois}. Cette correspondance a été utilisée la
première fois en fouille de données quand des algorithmes d'extraction des itemsets fermés fréquents ont été
proposés~\cite{pasquieretal99} et elle est aussi utilisée dans de nombreux travaux en apprentissage
conceptuel~\cite{wil92,mn00}. 

\noindent Étant donnée une base de données $\db$, les opérateurs $f$ et $g$ de Galois sont définis
par~:
\begin{itemize}
\item $f$, appelé {\em intension}, est une fonction de $2^\OO$ vers $2^\AA$ définie par
  $$f(O) = \set{a \in \AA \mid \forall o \in O,\; (a,o) \in \db },$$
\item $g$, appelé {\em extension},  est une fonction de $2^\AA$ vers $2^\OO$ définie par
  $$g(A) = \set{o \in \OO \mid \forall a \in A,\; (a,o) \in \db }.$$
\end{itemize}

Pour un ensemble $A$, $g(A)$ est aussi appelé \defnt{l'ensemble support} de $A$ dans $\db$. C'est 
l'ensemble des objets qui sont en relation avec tous les attributs de $A$. La \defnt{fréquence} de $A$
dans $\db$,
notée $\freq(A,\db)$ (ou plus simplement $\freq(A)$), est définie par $\freq(A)=\size{g(A)}$. 

Ces deux fonctions créent un lien entre l'espace des attributs et l'espace des objets. Pourtant,
comme les deux espaces n'ont a priori pas le même cardinal, aucune bijection n'est possible entre
eux. Cela signifie que plusieurs ensembles d'attributs ont la même image par $g$ dans l'espace des
objets et vice-versa. On peut donc définir 
deux relations d'équivalence $\ra$ et $\ro$ sur $2^\OO$ et $2^\AA$~:

\begin{itemize}
\item si $A$ et $B$ sont deux ensembles d'attributs, $A \ra B$ si $g(A)=g(B)$,
\item si $O$ et $P$ sont deux ensembles d'objets, $O \ro P$ si $f(O)=f(P)$. 
\end{itemize}

Dans chaque classe d'équivalence, il y a un élément particulier~: le plus grand élément d'une classe, au sens de
l'inclusion, est unique et appelé \defnt{ensemble d'attributs fermé} pour $\ra$ ou \defnt{ensemble d'objets fermé}
pour $\ro$. Les opérateurs $f$ et $g$ de Galois fournissent, par composition, deux opérateurs de
\defnt{fermeture} notés $h = f \circ g$ et $h' = g \circ f$. 
Les ensembles fermés sont les points fixes des opérateurs de fermeture et la fermeture
d'un ensemble est l'ensemble fermé de sa classe d'équivalence. Dans la suite, nous évoquerons
indifféremment $h$ ou $h'$ avec la notation $\cl$. 

\newcommand{\Aaabacad}{$\mathbf{\at 1\at 2\at 3\at 4}$}
\newcommand{\Abacad}{$\mathbf{\at 2\at 3\at 4}$}
\newcommand{\Abac}{$\mathbf{\at 2\at 3}$}
\newcommand{\Aaabac}{$\mathbf{\at 1\at 2\at 3}$}

\newcommand{\Aaad}{$\at 1\at 4$}
\newcommand{\Ac}{$\at 3$}
\newcommand{\Ab}{$\at 2$}
\newcommand{\Acad}{$\at 3\at 4$}
\newcommand{\Abad}{$\at 2\at 4$}
\newcommand{\Ad}{$\at 4$}
\newcommand{\Aaac}{$\at 1\at 3$}
\newcommand{\Aaab}{$\at 1\at 2$}
\newcommand{\Aa}{$\at 1$}
\newcommand{\Aaacad}{$\at 1\at 3\at 4$}
\newcommand{\Aaabad}{$\at 1\at 2\at 4$}

\newcommand{\Oaoboc}{$\mathbf{\ob 1\ob 2\ob 3}$}
\newcommand{\Oaob}{$\mathbf{\ob 1\ob 2}$}
\newcommand{\Oc}{$\mathbf{\ob 3}$}
\newcommand{\vide}{$\boldsymbol{\emptyset}$}

\newcommand{\Ob}{$\ob 2$}
\newcommand{\Oboc}{$\ob 2\ob 3$}
\newcommand{\Oa}{$\ob 1$}
\newcommand{\Oaoc}{$\ob 1\ob 3$}

\begin{figure}
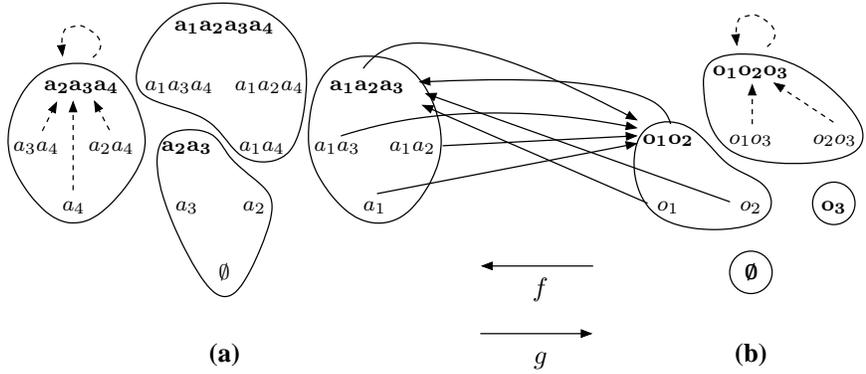

  \input treillis.pstex_t
  \caption{Les classes d'équivalence pour $\ra$ dans le treillis des attributs (a) et pour $\ro$ dans
  celui des objets (b). Les ensembles fermés sont en gras. Les flèches représentent les opérateurs $f$ et
  $g$ entre les classes de $\at{1}\at{2}\at{3}$ et $\ob{1}\ob{2}$. Les flèches en pointillés
  représentent les opérateurs de clôture $h$ et $h'$.} 
  \label{fig:treillis}
\end{figure}

Une paire $(A,O)$ constituée d'un ensemble d'attributs  fermé $A$ et de 
l'ensemble d'objets fermé correspondant $O$ est appelée un \defnt{concept formel}. 
L'ensemble des concepts de la base de
données $\db$ est noté~: 
$$\Concept(\db) = \set{(A,O)\mid f(O)=A \et g(A)=O}.$$

\begin{example}
  Dans la figure~\ref{fig:treillis}, les ensembles d'objets  fermés sont $\emptyset$, $\ob{3}$,
  $\ob{1}\ob{2}$, et $\ob{1}\ob{2}\ob{3}$. Les ensembles d'attributs fermés sont $\at{2}\at{3}$,
  $\at{2}\at{3}\at{4}$, $\at{1}\at{2}\at{3}$ et $\at{1}\at{2}\at{3}\at{4}$. Comme
  $g(\ob{1}\ob{2})=\at{1}\at{2}\at{3}$ et $f(\at{1}\at{2}\at{3})=\ob{1}\ob{2}$,
  $(\at{1}\at{2}\at{3},\ob{1}\ob{2})$ est un concept. Les autres concepts sont
  $(\at{2}\at{3},\ob{1}\ob{2}\ob{3})$, $(\at{2}\at{3}\at{4},\ob{3})$,
  $(\at{1}\at{2}\at{3}\at{4},\emptyset)$. 
\end{example}

\begin{property} \label{prop:decr} $A$ et $B$ sont des ensembles d'attributs, $O$ et $P$ des 
  ensembles d'objets et $E$ un ensemble d'attributs ou d'objets. 
  \begin{itemize}
  \item $f$ sont $g$ sont décroissantes par rapport à l'inclusion~: si $A \sseq B$ alors
  $g(B) \sseq g(A)$ et si $O \sseq P$, $f(P) \sseq f(O)$~;
  \item $f \circ g \circ f =f$~;
  \item $E$ est fermé si et seulement si $\cl(E) = E$ et sinon $E \sseq \cl(E)$~;
  \item $(A,O)$ est un concept si et seulement si $O$ est fermé et $A=f(O)$
  \end{itemize}
\end{property}

\subsection{Contraintes}
\label{sec:contraintes}

Afin de permettre au biologiste de focaliser son étude sur les concepts qui l'intéressent
réellement, nous lui laissons la possibilité de définir une contrainte qui devra être
satisfaite par tous les concepts extraits. 

\smallskip

Si on note $\BB$ l'ensemble des bases de données booléennes (i.e., des matrices booléennes),
on appelle \defnt{contrainte sur les concepts}
une fonction booléenne $\CC$ de $2^\AA \times 2^\OO \times \BB$.

Outre le fait qu'une contrainte permet de mieux cibler les ensembles extraits, leur utilisation,
lorsqu'elles sont efficacement intégrées à l'algorithme d'extraction, permet également de réduire
considérablement le temps de calcul. C'est ce qui explique l'intérêt croissant ces dernières années
pour l'étude des algorithmes d'extraction sous
contraintes.
Cependant, les contraintes utilisées dans ces algorithmes ne portent généralement que sur les
itemsets (et pas simultanément sur les itemsets et les ensembles d'objets).  Mais, dans la section
suivante, nous verrons comment projeter une contrainte sur les concepts pour obtenir une contrainte
ne portant plus que sur les objets, et ainsi pouvoir utiliser des techniques classiques d'extraction
sous contraintes (sauf que nous les utiliserons dans les données transposées). 

\smallskip

Parmi les contraintes portant sur les itemsets, la plus utilisée est sans doute la contrainte de
fréquence minimale $\Cfreq$.  Cette contrainte est satisfaite par les itemsets dont la fréquence est
supérieure à un seuil $gamma$ fixé par l'utilisateur : $\Cfreq(X)=(\freq(X) > \gamma)$.  On peut
également être intéressé par sa négation~: c'est-à-dire chercher des itemsets suffisamment rares et donc
utiliser une contrainte de fréquence maximale.  Il existe également de nombreuses contraintes
syntaxiques.  Une contrainte est syntaxique lorsqu'elle ne dépend pas de la matrice des données
$\db$. Par exemple, la 
contrainte\footnote{On notera $\CC(A)$ au lieu de $\CC(A,O,\db)$ lorsque l'expression de la contrainte
$\CC$ n'utilise pas $O$ et $\db$.} 
$\CC(A) = \at 1 \in A$ est syntaxique, alors que la contrainte de
fréquence ne l'est pas (en effet, la fréquence d'un itemset dépend des données).

Parmi les contraintes syntaxiques, les contraintes de ``sur-ensemble'' et de ``sous-ensemble''
permettent par combinaison (conjonction, disjonction, négation) de construire les autres contraintes
syntaxiques (cf. table~\ref{tab:contraintes}).  Étant donné un ensemble constant $E$, la contrainte
de sous-ensemble $\Csous E$ est définie par : $\Csous E(X)= (X \sseq E)$. La contrainte de
sur-ensemble $\Csur E$ est définie par : $\Csur E(X)= (X \supseteq E)$. Remarquons que comme nous
allons ensuite utiliser des contraintes sur les itemsets et les ensembles d'objets, les ensembles
$X$ et $E$ peuvent soit être tous les deux des itemsets soit tous les deux des ensembles d'objets. 

Lorsqu'une valeur numérique $a.v$ est associée à chaque attribut $a$ (par exemple un coût), on
peut définir d'autres contraintes syntaxiques du type~\cite{ngetal98} $\MAX(X) \comp \alpha$ (où
$\comp \in \set{<,>,\leq,\geq}$) pour différents opérateurs d'agrégation tels que $\MAX$, $\MIN$,
$\SOM$ (la somme), $\MOY$ (la moyenne).  Parmi ces contraintes, celles qui utilisent les opérateurs
$\MIN$ et $\MAX$ peuvent être récrites simplement en utilisant les contraintes $\Csur E$ et $\Csous
E$ en utilisant l'ensemble $\supalpha=\set{a\in \AA \mid a.v > \alpha}$ comme indiqué dans la
table~\ref{tab:contraintes}. 

\begin{table}[h]
  \hrule
  {\small
    \begin{align*}
      X \not \sseq E & \equiv \neg \Csous E (X)  & 
      \quad X \cap E = \emptyset &\equiv X \sseq \overline E\\ 
      X \not \supseteq E & \equiv \neg \Csur E(X)& 
      \quad X \cap E \neq \emptyset  &\equiv \neg (X \sseq \overline E)\\
      \MIN(X)> \alpha &\equiv X \sseq sup_\alpha  & 
      \quad \MAX(X) > \alpha &\equiv X \cap sup_\alpha \neq\emptyset\\
      \MIN(X)\leq\alpha &\equiv X \not\sseq sup_\alpha &
      \quad \MAX(X)\leq\alpha &\equiv X \cap sup_\alpha =\emptyset
    \end{align*}
    $$\size{X \cap E} \geq 2 \equiv   \bigvee_{1\leq i<j\leq n} e_ie_j \sseq X$$
  }
  \hrule
  \caption{Exemples de contraintes obtenues par combinaison des contraintes de sur-ensemble et 
    de sous-ensemble. 
    $E=\set{e_1,e_2,...,e_n}$ est un ensemble constant et $X$ un ensemble variable. Le complémentaire
    de $E$ dans $\AA$ ou dans $\OO$ est noté $\overline E$.} 
  \label{tab:contraintes}
\end{table}

Le fait de récrire toutes ces contraintes syntaxiques en utilisant uniquement les contraintes
$\Csous E$ et $\Csur E$ nous permettra de limiter le nombre de contraintes à étudier dans la
section~\ref{sec:proj} sur la projection des contraintes. 

Finalement, toutes ces contraintes peuvent être combinées pour construire une contrainte 
sur les concepts, par exemple $\CC(A,O) = (\at 1 \at 2 \sseq A  \et (O \cap \ob 4 \ob 5 = \emptyset))$. 

\smallskip

Pour pouvoir utiliser efficacement les contraintes dans les algorithmes d'extraction, il est
nécessaire d'étudier leurs propriétés. Ainsi, deux types de contraintes importantes ont été mises en
évidence : les contraintes monotones et les contraintes anti-monotones.  Une contrainte $\CC$ est
\defnt{anti-monotone} si $\forall A,B \; (A \sseq B \et \CC(B)) \Longrightarrow \CC(A)$.  $\CC$ est
\defnt{monotone} si $\forall A,B\; (A \sseq B \et \CC(A)) \Longrightarrow \CC(B)$. Dans les deux
définitions, $A$ et $B$ peuvent être des ensembles d'attributs ou d'objets. La contrainte de
fréquence est anti-monotone.  L'anti-monotonicité est une propriété importante, parce que les
algorithmes d'extraction par niveaux l'utilisent la plupart du temps pour élaguer l'espace de
recherche. En effet, quand un ensemble ne satisfait pas la contrainte, ses spécialisations non plus
et elles peuvent donc être élaguées~\cite{agrawaletal96}. 

Les compositions élémentaires de telles contraintes ont les mêmes propriétés~: la conjonction ou la
disjonction de deux contraintes anti-monotones (resp. monotones) est anti-monotone (resp. monotone). 
La négation d'une contrainte anti-monotone est monotone, et vice-versa. 

\subsection{Définition du problème}
\label{sec:def-pb}

Nous définissons la tâche d'extraction de concepts sous contraintes de la manière
suivante : étant donnés une base de données $\db$ et une contrainte $\CC$ sur 
les concepts, nous voulons extraire l'ensemble des concepts qui 
satisfont $\CC$, c'est-à-dire la collection
$\set{(A,O)\in \Concept(\db) \mid \CC(A,O,\db) }$.

\section{Projections de contraintes}
\label{sec:proj}

Pour extraire les concepts sous contraintes, nous proposons d'utiliser des techniques
classiques d'extraction de fermés sous contraintes dans la matrice transposée. 
Cependant, dans ces algorithmes, les contraintes possibles portent uniquement sur les itemsets. 
Donc, s'ils sont utilisés dans la matrice transposée, les contraintes porteront 
sur les ensembles d'objets. 
 
Pour permettre leur utilisation, nous allons donc étudier dans cette section un
mécanisme de ``projection'' de contrainte : étant donné une contrainte $\CC$ portant 
sur les concepts (c'est-à-dire à la fois sur l'intension et l'extension), nous voulons
calculer une contrainte $\pCC$ portant uniquement sur les ensembles d'objets et telle que
la collection des ensembles fermés d'objets satisfaisant cette contrainte soit exactement 
la collection des extensions des concepts satisfaisant $\CC$. De cette manière, 
La collection des concepts satisfaisant $\CC$ est exactement
la collection des $(f(O),O)$ tels que $O$ est fermé et satisfait la contrainte
projetée $\pCC$ :
\hspace{-3mm}
\begin{multline*}
  \set{ (A,O) \in \Concept(\db) \mid \CC(A,O,\db) } = \\
  \set{ (f(O),O)  \in \AA \times \OO \mid \pCC(O,\db) \et O \in \Ferme(\db)}. 
\end{multline*}
\hspace{-3mm}
Ainsi, pour résoudre notre problème, il suffira d'extraire les ensembles fermés 
d'objets $O$ satisfaisant $\pCC$ et de générer tous les concepts de la forme $(f(O),O)$. 

\subsection{Définitions et propriétés}
\label{sec:trans-def}

Cela signifie que nous voulons extraire des ensembles fermés d'objets $O$ tels que 
le concept $(f(O),O)$ satisfasse la contrainte $\CC$. 
Par conséquent, une définition naturelle de la projection de la contrainte $\CC$ est :

\begin{definition}[Contrainte projetée]\label{def:contrainte_proj}
  Étant donnée une contrainte $\CC$ sur les concepts, nous définissons la contrainte
  projetée de $\CC$ de la 
  façon suivante : $ \pCC(O,\db) = \CC(f(O),O,\db)$.
\end{definition}

La proposition suivante assure que l'on obtient bien le résultat voulu :

\begin{proposition}\label{prop:proj}
  Soit $\CC$ une contrainte sur les concepts, $\db$ une base de donnée et $\pCC$ la
  projection de la contrainte $\CC$.  Alors :
  \begin{multline*}
    \set{ (A,O) \in \Concept(\db) \mid \CC(A,O,\db) } = \\
    \set{ (f(O),O)  \in \AA \times \OO \mid \pCC(O,\db) \et O \in \Ferme(\db)}. 
  \end{multline*}
\end{proposition}

\begin{proof}
  $ (A,O) \in \Concept(\db) \et \CC(A,O,\db) \equ
     O \in \Ferme(\db) \et A=f(O) \et \CC(A,O,\db) \equ
     O \in \Ferme(\db) \et \CC(f(O),O,\db) \equ
     O \in \Ferme(\db) \et \pCC(O,\db).$
\end{proof}

\begin{example}
  Soit la contrainte $\CC(A)=(\at 4\not\in A)$. Sa projection est (par définition) :
  $\proj \CC(O)=(\at{4}\not\in f(O))$. Dans la matrice de la table~\ref{tab:database}, les ensembles
  fermés d'objets qui satisfont $\proj \CC$ sont $\ob{1}\ob{2}$ et $\ob{1}\ob{2}\ob{3}$. 
  Si on calcule les paires $(f(O),O)$ pour ces deux ensembles d'objets, on trouve :
  $(\at 1\at 2\at 3,\ob{1}\ob{2})$ et $(\at 2 \at 3,\ob{1}\ob{2}\ob{3})$ 
  qui sont bien les concepts satisfaisant $\CC$.
\end{example}

Par conséquent, pour extraire la collection des concepts qui satisfont $\CC$, nous pouvons utiliser
des algorithmes classiques d'extraction de fermés dans la matrice transposée avec la contrainte
$\pCC$.  Cependant, il faut vérifier que cette contrainte $\pCC$ est effectivement utilisable dans
ces algorithmes. 

Nous allons commencer par étudier les contraintes complexes, c'est-à-dire des contraintes
construites à partir de contraintes plus simples en utilisant des opérateurs booléens comme
la conjonction, la disjonction ou la négation. 

\begin{proposition}\label{prop:op_bool}
  Si $\CC$ et $\CC'$ sont deux contraintes sur les concepts, alors :\\
    \hspace*{\fill}$ \proj{\CC \et \CC'} = \pCC \et \proj{\CC'}$,\hfill 
    $\proj{\CC \ou \CC'} = \pCC \ou \proj{\CC'}$ et \hfill$\proj{\neg\CC} = \neg \pCC$.\hspace*{\fill} 

\end{proposition}

\begin{proof}
  Pour la conjonction : 
  $ \proj{\CC \et \CC'}(O) = (\CC \et \CC')(f(O),O)= \CC(f(O),O) \et \CC'(f(O),O)=(\pCC \et \proj{\CC'})(O)$. 
  La preuve est similaire pour la disjonction et la négation. 
\end{proof}

Cette proposition permet de ``pousser'' la projection dans les contraintes complexe. L'étape 
suivante est donc d'étudier ce qui se passe au niveau des contraintes élémentaires.

\begin{example}
  Si $\CC(A) = (\size A > 4 \et \freq(A) > 2) \ou ( A \cap \set{\at 1 \at 4} \neq \emptyset)$
  alors, d'après cette proposition, la projection $\proj \CC$ de $\CC$ est égale à 
$\proj \CC=( \proj{\CC_1} \et \proj{\CC_2}) \ou \proj{\CC_3}$ avec 
$\CC_1(A) = \size A > 4$,
$\CC_2(A) = \freq(A) > 2$ et
$\CC_3(A) = (A \cap \set{\at 1 \at 4} \neq \emptyset)$.
Nous verrons dans la section suivante comment calculer les projections de $\CC_1$, $\CC_2$ et $\CC_3$.
\end{example}

Ces contraintes élémentaires peuvent porter sur l'intension du concept (ex : $\CC(A,O)= ( \at
1 \in A )$) ou sur son extension (ex : $\CC(A,O)= ( \size{O \cap \ob 1 \ob 3 \ob 5} \geq 2$).
ou enfin sur les deux (Par exemple, la contrainte d'aire minimale sur les concepts : 
$\CC(A,O)= (\size A . \size O) > \alpha$). 
Les contraintes élémentaires qui ne portent que sur l'extension des concepts ne sont pas modifiées
par la projection, nous allons donc nous focaliser sur les contraintes portant sur les itemsets.

Les contraintes les plus efficacement prises en compte par les algorithmes d'extraction sous contrainte
sont les contraintes monotones et anti-monotones. Il est donc important d'étudier comment se comporte
la projection de contraintes par rapport à ces propriétés : 

\begin{proposition}\label{prop:proj-antimonotone}
Soit $\CC$ une contrainte sur les itemsets :
\begin{itemize}
\item si $\CC$ est anti-monotone alors $\pCC$ est monotone ;
\item si $\CC$ est monotone alors $\pCC$ est anti-monotone. 
\end{itemize}
\end{proposition}

\begin{proof}
  Si $O$ est un ensemble d'objet,  $\pCC(O)=\CC(f(O))$ par définition de la projection. 
  Or $f$ est décroissante par rapport à l'inclusion (cf. prop.~\ref{prop:decr}) d'où les propriétés. 
\end{proof}

\subsection{Projection de contraintes classiques}
\label{sec:contr-transp-de}

\begin{table}[tb]
  \begin{center}
    \begin{tabular}{cc}
      Contrainte $\CC(A)$ & Contrainte projetée $\pCC(O)$ \\\hline
      $\freq(A) \comp \alpha$ & $\size O \comp \alpha$\\
      $\size{A} \comp \alpha$ & $\freq(O) \comp \alpha$\\
      $A \sseq E$         & si $E$ est fermé : $g(E) \sseq O$\\
      & sinon : $O \not\sseq g(f_1) \et ... \et O \not\sseq g(f_m) $\\
      $E \sseq A$         & $O \sseq g(E)$\\
      $A \not\sseq E$     & si $E$ est fermé : $g(E) \not \sseq O$\\
      & sinon : $O \sseq g(f_1) \ou ... \ou O \sseq g(f_m) $\\
      $E \not\sseq A$     &  $O\not \sseq g(E)$\\
      $A\cap E = \emptyset$ & si $\overline E$ est fermé :  $ g(\overline E) \sseq O$\\
      & sinon : $O \not\sseq g(e_1) \et ... \et O \not\sseq g(e_n) $\\
      $A\cap E \neq \emptyset$ & si $\overline E$ est fermé : $ g(\overline E) \not\sseq O$\\
      & sinon : $O \sseq g(e_1) \ou ... \ou O \sseq g(e_n) $\\
      $\SOM(A) \comp \alpha$    & $\freq_p(O) \comp \alpha  $\\
      $\MOY(A) \comp \alpha$    & $\freq_p(O)/\freq(O) \comp \alpha $\\
      $\MIN(A) > \alpha$ & $\proj{A \sseq sup_\alpha}$ \\
      $\MIN(A) \leq \alpha$ & $\proj{A \not\sseq sup_\alpha}$ \\
      $\MAX(A) > \alpha$ & $\proj{A \cap sup_\alpha \neq\emptyset}$ \\
      $\MAX(A) \leq \alpha$ & $\proj{A \cap sup_\alpha =\emptyset}$ \\\hline
    \end{tabular}\\
    \smallskip
    $\comp \in \set{<,>,\leq,\geq}$
  \end{center}
  \caption{Contraintes projetées. $A$ est un ensemble variable d'attributs,  
    $E=\set{e_1,e_2,...,e_n}$ un ensemble fixé d'attributs,     
    $\overline E = \AA \setminus E=\set{f_1,f_2,...,f_m}$ son complémentaire et $O$ un ensemble 
    d'objets fermé.}
  \label{tab:proj_const}
\end{table}

Dans la section précédente, nous avons donné la définition de la projection de contrainte. Cette 
définition fait intervenir $f(O)$. Cela signifie que pour tester la contrainte projetée, il 
est nécessaire, pour chaque ensemble d'objets $O$, de calculer son intension $f(O)$. Certains
algorithmes, tels que \Charm~\cite{zaki02}, utilisent une structure de données 
particulière~--la représentation verticale des données--~et par conséquent calculent pour 
chaque ensemble $O$ l'ensemble $f(O)$. Cependant, beaucoup d'autres algorithmes 
n'utilisent pas cette structure et ne peuvent donc directement utiliser les contraintes 
projetées. C'est pour cette raison que dans cette section nous étudions les contraintes
projetées de contraintes classiques et nous calculons une expression de ces contraintes
ne faisant plus intervenir $f(O)$. 

\smallskip

Nous allons d'abord étudier la contrainte de fréquence minimale (qui est la contrainte la plus
courante) : $\Cfreq(A)=(\freq(A) > \gamma)$.  Par définition, sa contrainte projetée est :
$\proj{\Cfreq}(O)= (\freq(f(O)) > \gamma)$. Par définition de la fréquence,
$\freq(f(O))=\size{g(f(O))}=\size{\cl(O)}$ et si $O$ est un ensemble fermé d'objets, $\cl(O)=O$ et
par conséquent $\proj{\Cfreq}(O)= (\size O > \gamma)$.  Finalement, la projection de la contrainte
de fréquence minimale est une contrainte de taille minimale.  Si on avait considéré la contrainte de
fréquence maximale, on aurait évidement trouvé comme projection une contrainte de taille maximale.

De par la symétrie du problème, il découle que la projection de la contrainte de taille maximale
(resp. minimale) est la contrainte de fréquence : si $\CC(A)=(\size A \comp \alpha)$ alors
$\proj{\CC}(O)=( \size{f(O)} \comp \alpha )$. Or $\size{f(O)}$ est exactement la fréquence de $O$ si
on se place dans la matrice transposée.

\begin{sloppypar}
  Les deux propositions suivantes donnent l'expression de la projection des contraintes de sur-ensemble
et de sous-ensemble :
\end{sloppypar}

\begin{proposition}
  \label{prop:proj_sur-ens}
  Soit $E$ un itemset, alors :
  $$ \proj{\Csur E}(O) \equiv g(E) \supseteq \cl(O).$$
\end{proposition}
\begin{proof}
$\proj{\Csur E}(O) \equ (E \sseq f(O)) \Rightarrow (g(E) \supseteq g\circ f(O)) \equ 
(g(E) \supseteq \cl(O))$.  Réciproquement, $(g\circ f(O) \sseq g(E)) 
\Rightarrow (f\circ g\circ f(O) \supseteq f\circ g(E)) 
\Rightarrow (f(O) \supseteq \cl(E)) \Rightarrow f(O) \supseteq E.$ 
\end{proof}

\begin{proposition}
  \label{prop:proj_sous-ens}
  Soit $E$ un itemset, alors, si $E$ est fermé :
  $$ \proj{\Csous E}(O) \equiv  g(E) \sseq \cl(O),$$
  si $E$ n'est pas fermé, on pose $\overline E = \AA \setminus E = \set{f_1, ..., f_m}$ et : 
  $$ \proj{\Csous E}(O) \equiv (\cl(O) \not\sseq g(f_1) \et \cl(O) \not\sseq g(f_2) \et ... \et 
  \cl(O) \not\sseq g(f_m).$$
\end{proposition}

\begin{proof}
  $\proj{\Csous E}(O) \equ \CC_{\sseq E}(f(O)) \equ (f(O) \sseq E) \Rightarrow (g\circ f(O)
  \supseteq g(E)) \equ (\cl(O) \supseteq g(E))$.  Réciproquement, (si $E$ est fermé): $(g(E) \sseq g
  \circ f(O)) \Rightarrow (f\circ g(E) \supseteq f\circ g\circ f(O)) \Rightarrow (\cl(E) \supseteq
  f(O)) \Rightarrow (E \supseteq f(O))$. Si $E$ n'est pas fermé, on récrit la contrainte : $(A \sseq
  E) = f_1 \not\in A \et ... \et  f_m \not\in A$ et on utilise les propositions
  \ref{prop:op_bool}~et~\ref{prop:proj_sur-ens}. 
\end{proof}

La table~\ref{tab:proj_const} récapitule les contraintes projetées de contraintes classiques. Les
contraintes de fréquence et de taille ont été traitées plus haut. Les deux propriétés précédentes,
avec l'aide de la table~\ref{tab:contraintes} et de la proposition~\ref{prop:op_bool} nous
permettent de calculer la projection des contraintes syntaxiques, exceptées les contraintes
utilisant les opérateurs d'agrégation $\MOY$ et $\SOM$.  Dans cette table, on suppose que l'ensemble
d'objets $O$ est fermé.  Cela n'est pas une restriction importante dans la mesure où nous ne nous
intéressons qu'à des algorithmes d'extraction de fermés (ces fermés serviront à générer les
concepts).

Examinons maintenant les contraintes utilisant les opérateurs d'agrégation $\MOY$ et $\SOM$.  Par
définition, les contraintes projetées sont : $\MOY(f(O)) \comp \alpha$ et $\SOM(f(O)) \comp \alpha$.
Il faut donc trouver une expression de $\MOY(f(O))$ et $\SOM(f(O))$ ne faisant plus intervenir $f$.
En fait, il suffit d'étudier l'opérateur $\SOM$ car $\MOY(f(O)) = \SOM(f(O))/\size{f(O)} =
\SOM(f(O))/\freq(O)$ donc si nous trouvons une expression de $\SOM(f(O))$ dans la base projetée,
nous obtiendrons aussi une expression pour $\MOY(f(O))$.

L'ensemble $f(O)$ est un ensemble d'attribut, donc dans la matrice transposée, c'est un ensemble de
lignes.  Les valeurs $a.v$ sur lesquelles la somme est calculée sont attachées aux attributs $a$ et
donc aux lignes de la matrice transposée. La valeur $\SOM(f(O))$ est donc la somme de ces valeurs
$v$ sur toutes les lignes de $f(O)$, c'est-à-dire les lignes contenant $O$. Autrement dit,
$\SOM(f(O))$ est une fréquence pondérée par les valeurs $v$ (nous notons cette fréquence pondérée
$\freq_p$). Celle-ci peut être facilement calculée par les algorithmes en plus de la fréquence
``classique'' $\freq$. Il suffit pour cela, lors de la passe sur les données, d'incrémenter cette
fréquence pondérée de $a.v$ pour chaque ligne $a$ contenant $O$.

\smallskip

Ces expressions de la contrainte projetée sont intéressantes car elles n'impliquent plus le calcul
de $f(O)$ pour chaque ensemble devant être testé. Les ensembles $g(\overline E)$ or $g(e_i)$ qui
apparaissent dans ces contraintes peuvent quant à eux être calculés une fois pour toute lors de la
première passe sur les données (en effet, l'ensemble $E$ est constant).

\begin{example}
  Considérons la contrainte $\CC_3(A) = (A \cap \set{\at 1 \at 4} \neq \emptyset)$ de l'exemple précédent.
  Dans la table~\ref{tab:database}, l'itemset
  $\overline{\at{1}\at{4}}=\at{2}\at{3}$ est fermé. Par conséquent, la contrainte projetée est
  $\proj{\CC_3}(O)=(g(\at{2}\at{3}) \not\sseq O)$.  Comme
  $g(\at{2}\at{3})=\ob{1}\ob{2}\ob{3}$, $\proj{\CC_3}(O)=(\ob{1}\ob{2}\ob{3} \not\sseq O)$. 
  La projection de la contrainte  
  $\CC(A) = (\size A > 4 \et \freq(A) > 2) \ou ( A \cap \set{\at 1 \at 4} \neq \emptyset)$
   de l'exemple~3 est donc :
  $\proj \CC(O) = ( \freq(O) > 4 \et \size O > 2) \ou (\ob{1}\ob{2}\ob{3} \not\sseq O)$.
\end{example}

\section{Utilisation de la projection de contraintes}
\label{sec:util_proj}

Dans cette section, nous présentons deux stratégies pour extraire les concepts satisfaisant une
contrainte $\CC$ et ainsi résoudre le problème posé dans la section~\ref{sec:def-pb}. 

La première stratégie utilise les algorithmes classiques d'extraction de fermés :
\begin{enumerate}
\item Calculer la contrainte projetée $\pCC$ de $\CC$ en utilisant la table~\ref{tab:proj_const} et
  la propriété~\ref{prop:op_bool} ;
\item Utiliser un algorithme pour l'extraction de fermés sous contraintes dans la matrice transposée
  (comme par exemple, ceux proposés dans \cite{bonchi04a} ou \cite{boulicautjeudy01a}) avec la
  contrainte $\pCC$.  Il est aussi possible d'utiliser des algorithmes d'extraction de fermés
  fréquent tels que \Charm~\cite{zaki02}, \Carpenter~\cite{pct03} ou
  \Closet~\cite{peietal00b} en leur rajoutant une étape d'élagage supplémentaire pour
  traiter la contrainte (à la manière de ce qui est fait dans~\cite{hankdd00}). 
\item Ces algorithmes extraient des ensembles fermés. Cela signifie qu'ils vont retourner les
  ensembles d'objets fermés (car nous travaillons dans la matrice transposée) qui satisfont la
  contrainte $\pCC$. Il faut alors pour chacun de ces fermés calculer son intension $f(O)$, d'après
  la proposition~\ref{prop:proj}, les paires $(f(O),O)$ ainsi formées seront exactement les
  concepts qui satisfont la contrainte $\CC$.  Le calcul de $f(O)$ peut être fait lors d'une
  dernière passe sur les données ou alors intégré dans les algorithmes. En fait, ces algorithmes
  calculent les intensions lors du calcul de la fréquence des ensembles (la fréquence de $O$ est
  $\size{f(O)}$).  Il suffit donc de les modifier pour qu'ils stockent ces intensions. 
\end{enumerate}

\begin{example}
  Imaginons que nous voulions extraire les concepts satisfaisant la contrainte $\CC(A)= (A \cap
  \set{\at 1 \at 4} \neq \emptyset)$ avec cette stratégie. La projection de $\CC$ est
  (cf.~exemple~4) : $\pCC(O)=(\ob{1}\ob{2}\ob{3} \not\sseq O)$. Les ensembles fermés d'objets qui
  satisfont cette contrainte sont $T=\set{\emptyset,\ob{1}\ob{2},\ob{3}}$ (calculés dans la matrice
  transposée avec un algorithme d'extraction de fermés sous contraintes).  Nous pouvons ensuite
  calculer les concepts correspondants qui sont : $(\at 1 \at 2 \at 3 \at 4,\emptyset)$, $(\at 1 \at
  2 \at 3,\ob{1}\ob{2})$ et $(\at 2 \at 3 \at 4,\ob{3})$.
\end{example}

La seconde stratégie est basée sur le nouvel algorithme \Dminer~\cite{brb04}. Cet
algorithme extrait des concepts sous une contrainte $\CC$ qui est la conjonction d'une contrainte
monotone sur les attributs et d'une contrainte monotone sur les objets. Il ne peut cependant pas
traiter le cas où des contraintes anti-monotones sont utilisées. 

Notre stratégie consiste alors à projeter les contraintes anti-monotones définies dans l'espace des
attributs sur l'espace des objets et à projeter les contraintes anti-monotones définies dans
l'espace des objets sur l'espace des attributs. En effet, d'après la
proposition~\ref{prop:proj-antimonotone}, la projection transforme une contrainte anti-monotone en
une contrainte monotone. Cela permet donc d'utiliser {\Dminer} avec des contraintes
monotones et anti-monotones. Nous n'avons présenté que la projection des contraintes de l'espace des
attributs sur l'espace des objets. Cependant, la projection dans l'autre sens est similaire. En
fait, il suffit de remplacer la fonction $f$ par la fonction $g$. 

\section{Conclusion}
\label{sec:conclusion}

L'analyse des données d'expression de gènes pose un problème spécifique pour l'extraction de
motifs~: les données contiennent beaucoup plus de colonnes que de lignes, ce qui rend les 
algorithmes d'extraction classiques inopérants. Dans ce cas, extraire les motifs dans 
la matrice transposée permet de s'affranchir de ce problème. 

La transposition a déjà été étudiée dans le cas de la contrainte de fréquence, mais l'étude générale
de ce qui se passe dans le cas d'une contrainte complexe restait à faire. Cette étude nous a permis
de proposer des stratégies pour extraire des concepts sous contraintes. Ces stratégies, plutôt que
de proposer un nouvel algorithme, se fondent sur l'utilisation d'algorithmes classiques et éprouvés
d'extraction de fermés ou de concepts. Afin de rendre leur utilisation possible, nous avons défini
une opération de projection des contraintes et nous avons étudié ses propriétés ainsi que les
projections de contraintes classiques.

\end{document}